\documentclass{article}
\usepackage{spconf,amsmath,graphicx}

\usepackage{times}
\usepackage{latexsym}
\usepackage{amsfonts}
\usepackage{enumitem}
\usepackage{booktabs}
\usepackage{threeparttable}
\usepackage{mathrsfs}
\usepackage{latexsym}
\usepackage{stmaryrd}
\usepackage{multicol}
\usepackage{multirow}
\usepackage{xcolor}
\usepackage{bbding}
\usepackage{makecell}
\usepackage{arydshln}
\usepackage{tabularx}
\usepackage[T1]{fontenc}

\usepackage[utf8]{inputenc}

\usepackage{microtype}

\usepackage{inconsolata}

\usepackage{cleveref}
\crefname{section}{§}{§§}
\Crefname{section}{§}{§§}

\definecolor{domaincolor}{RGB}{251, 231, 163}
\definecolor{intentcolor}{RGB}{194, 214, 236}
\definecolor{slotcolor}{RGB}{241, 205, 177}
\definecolor{wordcolor}{RGB}{242, 242, 242}

\title{A BiRGAT Model for Multi-intent Spoken Language Understanding with Hierarchical Semantic Frames}
    \name{Hongshen Xu$^{1*}$, Ruisheng Cao$^{1*}$, Su Zhu$^{2\dagger}$, Sheng Jiang$^{2}$, Hanchong Zhang$^{1}$, Lu Chen$^{1}$, Kai Yu$^{1\dagger}$ \thanks{* Equal Contribution.} \thanks{$\dagger$ Su Zhu and Kai Yu are the corresponding authors.}}

\address{   $^{1}$MoE Key Lab of Artificial Intelligence, AI Institute \\X-LANCE Lab, Department of Computer Science and Engineering\\ Shanghai Jiao Tong University, Shanghai, China\\
$^{2}$AISpeech Co., Ltd., Suzhou, China
}
  
\begin{document}
\ninept
    \maketitle
\begin{abstract}
Previous work on spoken language understanding~(SLU) mainly focuses on single-intent settings, where each input utterance merely contains one user intent. This configuration significantly limits the surface form of user utterances and the capacity of output semantics. In this work, we firstly propose a \underline{M}ulti-\underline{I}ntent dataset which is collected from a realistic in-\underline{V}ehicle dialogue \underline{S}ystem, called MIVS. The target semantic frame is organized in a $3$-layer hierarchical structure to tackle the alignment and assignment problems in multi-intent cases. Accordingly, we devise a BiRGAT model to encode the hierarchy of ontology items, the backbone of which is a dual relational graph attention network. Coupled with the $3$-way pointer-generator decoder, our method outperforms traditional sequence labeling and classification-based schemes by a large margin.
Ablation study in transfer learning settings further uncovers the poor generalizability of current models in multi-intent cases.
\end{abstract}
\section{Introduction}
Spoken language understanding~(SLU,~\cite{slu}), which aims to parse the user utterance into a semantic frame, plays a critical role in building dialogue systems. Previous works focus on parsing utterances containing merely one intent. This simplification decomposes the original task into two sub-tasks, namely slot filling and intent detection~\cite{tur2011spoken}. When it comes to multi-intent cases~\cite{agif}, the traditional sequence labeling~\cite{zhu2017encoder}~(for slot filling) and sentence classification~\cite{joint-mi}~(for intent detection) schemes are not applicable due to 1) the slot-value alignment problem, and 2) the slot-intent assignment issue.

\begin{figure}[htbp]
    \centering
    \includegraphics[width=0.48\textwidth]{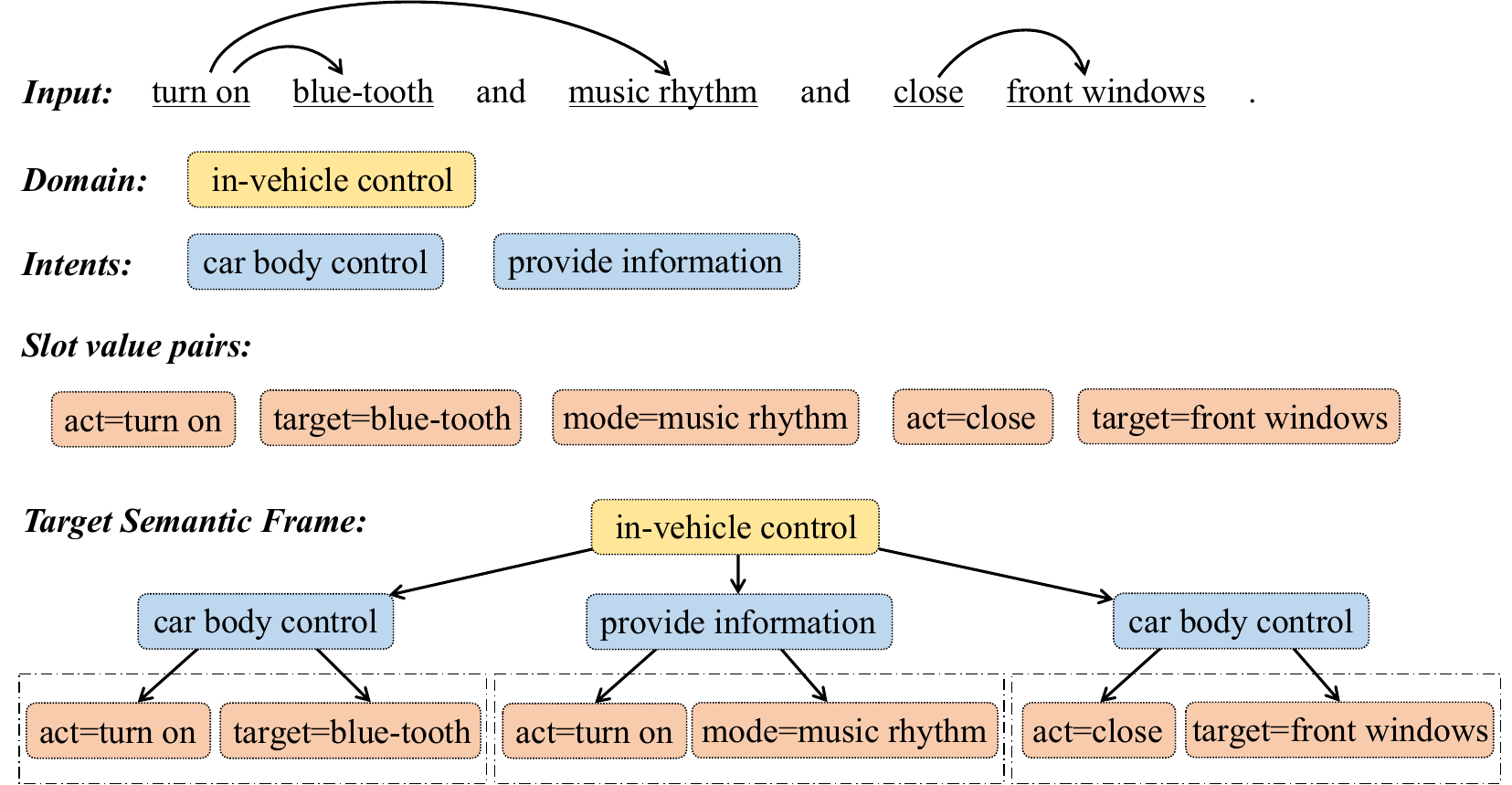}
    \caption{A multi-intent example from MIVS dataset.}
    \label{fig:intro}
    \vspace{-4mm}
\end{figure}
Firstly, the same slot value may be aligned to multiple slots or used several times in the target semantic representation~(duplicate alignments). As shown in Figure~\ref{fig:intro}, the slot-value pair ``{\tt act=turn on}" is used twice to control the ``{\tt blue-tooth}" and ``{\tt music rhythm}". Besides, some frequently used slot values may be implicitly mentioned and will not occur as a continuous span in the input utterance, which is also known as the \emph{unaligned slot value} problem~\cite{unaligned}. In both cases, the traditional sequence labeling strategy is not applicable.

Furthermore, in multi-intent cases, if slot filling and intent detection are treated as separate tasks, the affiliation relationship from slot to intent can not be determined. In other words, slot-value pairs need to be clustered and allocated to their parent intent, restricted by the hierarchy of ontology items. As illustrated in Figure~\ref{fig:intro}, we need to distinguish that the ``{\tt blue-tooth}" and ``{\tt front windows}" should be ``{\tt turned on} and ``{\tt closed}" respectively. And a simple modification of multi-class intent detection into multi-label classification~\cite{co-guiding,group} will lose this slot-intent assignment information.

To this end, we firstly construct a large-scale \underline{M}ulti-\underline{I}ntent Chinese dataset collected from a realistic in-\underline{V}ehicle \underline{S}ystem~(MIVS) with $105,240$ data points. It also contains 
multi-domain samples where each input utterance involves two domains since users often lazily make their requests all at once for convenience. The target semantic frame is organized as a 3-layer tree, rooting from domains to intents and then slots~(exemplified in the lower part of Figure~\ref{fig:intro}). In accordance with this structured representation, we inject the hierarchy knowledge of ontology items into the encoder through two dual relational graph attention networks~(RGAT,~\cite{rgat}). As for the decoder, after linearizing the tree representation into a string sequence with sentinal tokens, an adapted pointer-generator auto-regressive network~\cite{ptrgen} is utilized to selectively copy raw question words and ontology items to the output side. Experiments on two multi-intent datasets with hierarchical semantics, Chinese MIVS~(this work) and English TOPv2~\cite{topv2}, demonstrate the advantage of our proposed BiRGAT framework over traditional methods. Codes and data are publicly available~\footnote{https://github.com/importpandas/MIVS\_BIRGAT}.

\section{Dataset Construction}
\label{sec:dataset}
Given the ontology $O=\{o_i\}_{i=1}^{|O|}$ where $o_i$ denotes a domain, intent, or slot, SLU converts an utterance $Q=(q_1,q_2,\cdots,q_{|Q|})$ into the semantic frame $y$. The hierarchical structure of domain$\rightarrow$intent$\rightarrow$slot is also provided as input structural priors.

\begin{figure*}[htbp]
    \centering
    \includegraphics[width=0.95\textwidth]{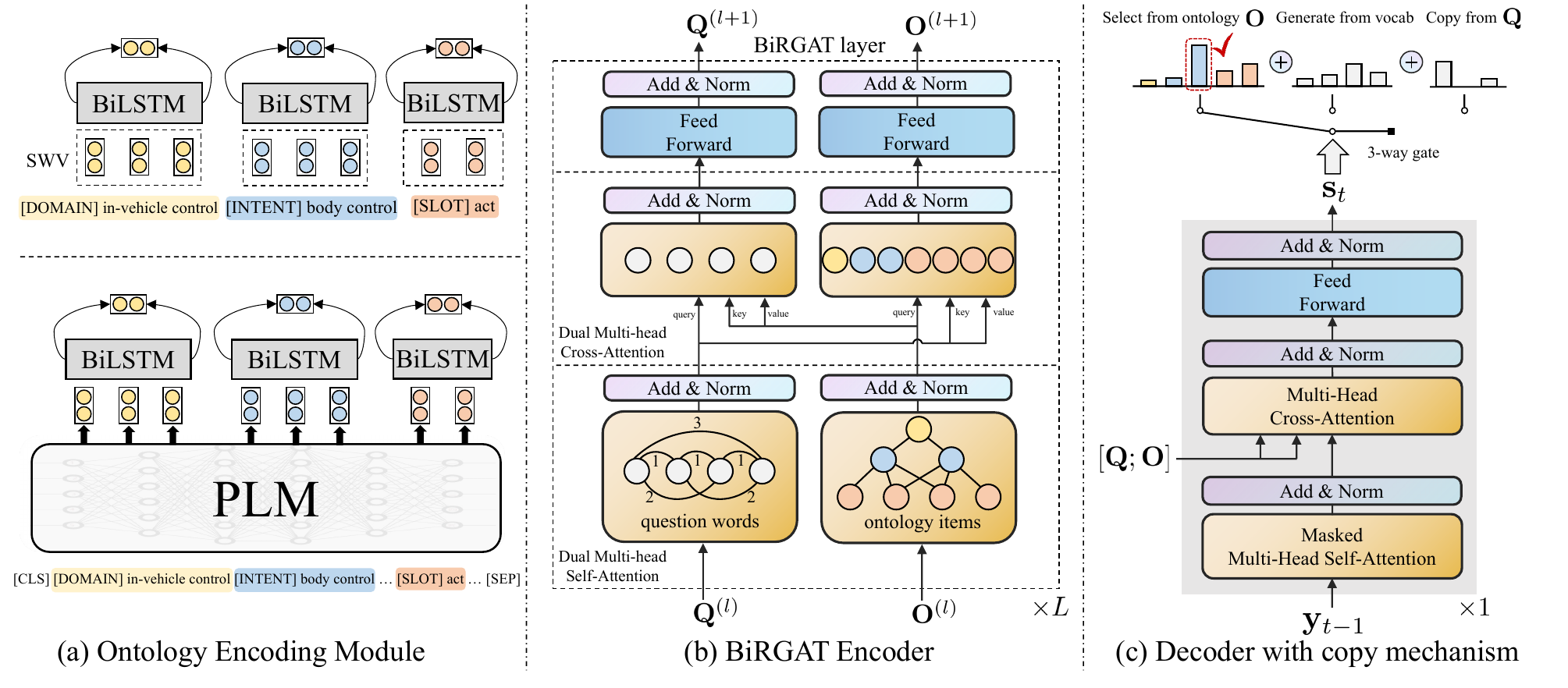}
    \caption{An overview of the BiRGAT model architecture.}
    \label{fig:model}
\end{figure*}

Prevalent benchmark ATIS~\cite{atis} or SNIPS~\cite{snips} simplifies the task by focusing on one single domain, considering one intent per utterance, and ignoring the hierarchy. In this work, we generalize to more practical scenarios where each utterance involves multiple intents and even multiple domains. 
Accordingly, the target semantic frame $y$ is labeled as a tree to reflect the structure. 
The comparison to previous benchmarks is present in Table~\ref{tab:comparison}.

\begin{table}[htbp]
  \centering
  \resizebox{0.49\textwidth}{!}{
    \begin{tabular}{c|cccc}
    \hline
    \hline
    \textbf{Dataset} & \makecell[c]{\bf Multi-\\\bf domain} & \makecell[c]{\bf Multi-\\\bf intent} & \makecell[c]{\bf Hierarchical\\\bf Annotation} &\makecell[c]{\bf \# of\\\bf Samples}\\
    \hline\hline
    ATIS  & \XSolidBrush & \XSolidBrush & \XSolidBrush & $6k$ \\
    SNIPS & \XSolidBrush & \XSolidBrush & \XSolidBrush & $14k$ \\
    MixATIS & \XSolidBrush & {\large\checkmark} & \XSolidBrush & $20k$ \\
    MixSNIPS & \XSolidBrush & {\large\checkmark} & \XSolidBrush & $50k$ \\
    TOPv2 & \XSolidBrush & {\large\checkmark} & {\large\checkmark} & $181k$ \\
    \textbf{MIVS~(ours)}  & {\large\checkmark} & {\large\checkmark} & {\large\checkmark} & $105k$ \\
    \hline
    \hline
    \end{tabular}%
}
  \caption{Comparison to previous benchmarks.}
  \label{tab:comparison}%
\end{table}%

The multi-intent MIVS dataset contains $5$ different domains, namely {\it map}, {\it weather}, {\it phone}, {\it in-vehicle control} and {\it music}. The dataset can be split into two parts: \textbf{single-domain} examples contain both single-intent and multi-intent cases, which are collected and manually annotated from a realistic industrial in-vehicle environment; \textbf{cross-domain} examples are automatically synthesized following MixATIS~\cite{agif}. Concretely, we extract two utterances from two different domains and concatenate them by conjunction words such as ``and''.
The semantic tree is serialized as an output token sequence by inserting sentinel tokens such as brackets for clustering.

\section{Model Architecture}
The entire BiRGAT model can be split into three parts as shown in Figure~\ref{fig:model}. Firstly, we adopt an ontology encoding module to obtain the initial ontology embedding~(\cref{sec:ont_enc}). Next, these features are further encoded via a dual RGAT for structural knowledge enhancement   ~(\cref{sec:dual_rgat}). After that, an auto-regressive decoder is employed to produce the serialized semantic frame based on the encoded memory~(\cref{sec:dec}).

\subsection{Ontology Encoding Module}\label{sec:ont_enc}
Inspired by the concept of label embeddings~\cite{label}, given an ontology item $o_i^d\in O^d=\{o^d_i\}_{i=1}^{|O^d|}$ from one specific domain $d$~(e.g., music), we can initialize its embedding $\mathbf{o}^d_i\in\mathbb{R}^{1\times m}$~(dimension is $m$) from either static word vectors~(SWV) or pre-trained language models~(PLMs) such as BERT~\cite{bert}, see Figure~\ref{fig:model}(a).

Concretely, we first concatenate all ontology items $o^d_i=(o^d_{i1}, \cdots, o^d_{il})$ as well as their semantic type $o^d_{i0}\in\{\textsc{Domain},\textsc{Intent},
\\ \textsc{Slot}\}$ to form a unified ontology sequence. Then the input sequence is fed into either a SWV module or PLM module to get token-level ontology embeddings. Finally, the forward and backward hidden states from a type-aware single-layer Bi-LSTM are concatenated as the ontology embedding $\mathbf{o}^d_i\in\mathbb{R}^{1\times m}$ for each ontology item $o^d_i$.


For domain $d$, we stack the initial embeddings of all ontology items to attain the domain feature matrix $\mathbf{O}^d\in\mathbb{R}^{|\mathrm{O}^d|\times m}$. In multi-domain cases, we take one more step to stack matrices $\mathbf{O}^d$ from all possible domains $d$ and get the entire matrix $\mathbf{O}^{(0)}\in\mathbb{R}^{|\mathrm{O}|\times m}$, where $\mathrm{O}=\bigcup_d\mathrm{O}^d$. Otherwise, $\mathbf{O}^d$ directly serves as $\mathbf{O}^{(0)}$.

For input question $Q=(q_1,q_2,\cdots,q_{|\mathrm{Q}|})$, the initial features $\mathbf{Q}^{(0)}\in\mathbb{R}^{|\mathrm{Q}|\times m}$ can also be initialized from SWV or PLM.

\subsection{BiRGAT Encoder}
\label{sec:dual_rgat}
After obtaining the initial matrix $\mathbf{Q}^{(0)}$ and $\mathbf{O}^{(0)}$ of question words and ontology items, this module further enriches features with structural knowledge and cross-segment information. The BiRGAT encoder consists of $L$ layers, the computation for layer $l$ is
$$\mathbf{Q}^{(l+1)},\mathbf{O}^{(l+1)}=\text{BiRGAT}(\mathbf{Q}^{(l)}, \mathbf{O}^{(l)}),$$
where $0\le l\le L-1$. Each layer includes three sub-modules, namely 1) dual multi-head self-attention, 2) dual multi-head cross-attention, and 3) feedforward network. Each sub-module is also wrapped with residual connections and LayerNorm function.

\subsubsection{Dual Multi-head Self-attention}
Transformer~\cite{transformer} architecture is a specific implementation of graph attention network~(GAT,~\cite{gat}). To integrate the hierarchical structure among ontology items, we adapt the multi-head self-attention module by relative position embeddings~\cite{rpe}. Concretely, edge feature $\mathbf{z}_{ij}$ is introduced from adjacent ontology item $o_j$ to $o_i$ when computing the attention weight $e_{ij}$ and attention vector $\tilde{\mathbf{o}}_i$,
\begin{align*}
e_{ij}&=\frac{(\mathbf{o}_i\mathbf{W}_q)(\mathbf{o}_j\mathbf{W}_{k}+\mathbf{z}_{ij}\mathbf{W}_z)^{\mathrm{T}}}{\sqrt{m}},\\
\mathbf{\tilde{o}}_i&=\sum_{j\in\mathcal{N}(i)}a_{ij}(\mathbf{o}_j\mathbf{W}_v+\mathbf{z}_{ij}\mathbf{W}_z),
\end{align*}
where $a_{ij}$ is the softmax version of $e_{ij}$ and $\mathcal{N}(i)$ denotes the neighborhood of $o_i$. 

We design the relation type $z_{ij}$ between ontology items mainly to address the multi-intent and multi-domain problem.
The slot items only have \textit{slot-intent} relations with their parent intents. Similarly, the slot and intent items only have \textit{slot-domain} and \textit{intent-domain} relations with their parent domains, respectively. This design not only models the hierarchical semantic frame of the ontology structure but also mitigates the inter-domain information interference.

As for the question, we construct a complete graph among question words and utilize relative distances between words as the relation $z_{ij}$. To avoid over-parametrization, all edge features $\mathbf{z}_{ij}$ are shared across different layers and heads.

\subsubsection{Dual Multi-head Cross-attention}
This sub-module aims to gather information for each question word from the counterpart ontology items~(and vice versa). By analogy to the multi-head cross-attention module in Transformer decoder, features of ontology items are incorporated as key/value vectors to enrich the representation of each question word. The symmetric part can be easily inferred, see the middle part of Figure~\ref{fig:model}(b).

After the relational graph encoding and cross-segment encoding, features of question words and ontology items are passed into a feedforward network. The outputs $\mathbf{Q}^{(L)}$ and $\mathbf{O}^{(L)}$ of the last layer $L$ serve as the final encoder memory $\mathbf{Q}$ and $\mathbf{O}$.

\subsection{Decoder with Copy Mechanism}
\label{sec:dec}
Given encoder memory $\mathbf{Q}$ and $\mathbf{O}$, the output token sequence $y=(y_1,y_2,\cdots,y_{|y|})$ is produced auto-regressively via a single-layer Transformer decoder. The decoder hidden state $\mathbf{s}_t$ at timestep $t$ is
$$\mathbf{s}_t=\text{TransformerDecoder}(\mathbf{y}_{<t}, [\mathbf{Q};\mathbf{O}]).$$
Notice that $y_t$ can be words in slot values, sentinel tokens~(e.g., brackets), or an ontology item in the pre-defined specification. It is difficult to generate an ontology item token-by-token because a simple morphological change or synonym substitution will cause parsing errors while post-processing the linearized semantic frame $y$. Thus, we introduce a three-way gate to control the action of generating a token from a fixed vocabulary, copying a word from the question memory $\mathbf{Q}$, and selecting an ontology item from the ontology memory $\mathbf{O}$. Formally, given the decoder hidden state $\mathbf{s}_t\in\mathbb{R}^{1\times m}$,
\begin{align*}
\mathbf{g}_t&=\text{softmax}(\mathbf{s}_t\mathbf{W}_g),\quad \mathbf{g}_t\in\mathbb{R}^{1\times 3},\\
P_{\text{gen}}(w_i)&=\text{softmax}_i(\mathbf{s}_t\mathbf{W}_{\text{gen}}\phi(w_i)^{\mathrm{T}}),\\
P_{\text{copy}}(w_i)&=\sum_{k:\ q_k=w_i}\text{PtrNet}(\mathbf{s}_t, \mathbf{Q})[k],\\
P_{\text{select}}(o_i)&=\text{PtrNet}(\mathbf{s}_t, \mathbf{O})[i],\\
P(y_t|\mathbf{s}_t,\mathbf{Q},\mathbf{O})&=g_{t1}P_{\text{gen}} + g_{t2}P_{\text{copy}}+g_{t3}P_{\text{select}},
\end{align*}
where $\phi(w_i)$ returns the word embedding of $w_i$ in a fixed vocabulary which is shared with the encoder, and $\text{PtrNet}(\mathbf{s}_t, \mathbf{Q})[k]$ denotes the probability of choosing the $k$-th entry~(row) in memory $\mathbf{Q}$ which is implemented as the average weight from different heads of a multi-head cross-attention module~(known as the pointer network,~\cite{ptrgen}).
The training objective is decoupled as
$$\mathcal{L}=-\sum_{t=1}^{|y|} \log P(y_t|y_{<t},\mathbf{Q},\mathbf{O}).$$
\section{Experiment}
\label{sec:exp}
\subsection{Datasets}
We experiment on two multi-intent SLU datasets, namely Chinese MIVS~(this work) and English TOPv2~\cite{topv2}. The original output format of TOPv2 is inconsistent with our annotation. Thus, we convert the output labels of TOPv2 into our $3$-layer hierarchical structure~(\cref{sec:dataset}). We report the sentence-level accuracy as the evaluation metric.

\subsection{Implementation Details}
Our model is implemented with Pytorch and {\tt transformers} library. The hidden dimension $m$ is $256/512$ for SWV and PLM respectively. The number of layers for the BiRGAT encoder is $2$. As for the pointer-generator decoder, the number of layers is fixed to $1$. The number of heads and dropout rate are set to $8$ and $0.2$ respectively. We use AdamW~\cite{adam} optimizer with a linear warmup scheduler. The warmup ratio of total training steps is $0.1$.
The leaning rate and weight decay rate are $5e\textrm{-}4/1e\textrm{-}4$ for SWV and $2e\textrm{-}4/0.1$ for PLM. We train all the models with a batch size of $20$ and $100k$ training iterations. For inference, we adopt beam search with size $5$.

\subsection{Main Results}\label{sec:main_res}
\begin{table}[htbp]
  \centering
  \resizebox{0.48\textwidth}{!}{
    \begin{tabular}{l|l|cc|cc}
    \hline

    \hline
    \multirow{2}{*}{\textbf{Init}} & \multirow{2}{*}{\textbf{Method}} & \multicolumn{2}{c|}{\textbf{MIVS}} & \multicolumn{2}{c}{\textbf{TOPv2}} \\
\cline{3-6}          &       & \textbf{Dev} & \textbf{Test} & \textbf{Dev} & \textbf{Test} \\
    \hline\hline
    \multirow{4}{*}{SWV} & SL    & 14.3  & 14.2  & 28.1  & 28.9  \\
          & \qquad +CLF & 21.2  & 21.4  & 35.6  & 36.3  \\
    \cdashline{2-6}
          & BiRGAT & \textbf{84.9}  & \textbf{85.6}  & \textbf{86.1}  & \textbf{85.9}  \\
          & \qquad w/o Copy & 72.5  & 72.8  & 82.0  & 82.1  \\
    \hline
    \multicolumn{2}{l|}{BART} & 27.0  & 26.8  & 84.3  & 83.7  \\
    \multicolumn{2}{l|}{\qquad w/ Copy} & 63.7  & 63.3  & 87.6  & 87.2  \\
    \hline
    \multirow{4}{*}{BERT} & SL & 14.8 & 14.9 & 29.1  &  29.8 \\
    & \qquad +CLF & 23.0 & 23.1 & 36.8 &  37.5 \\
    \cdashline{2-6}
    & BiRGAT  & \textbf{89.3}  & \textbf{89.3}  & \textbf{88.0}  & \textbf{87.8}  \\
    & \qquad w/o Copy & 75.6  & 75.5  &  84.1 &  84.1   \\
    \hline
    RoBERTa & \multirow{2}{*}{BiRGAT} & 89.3  & 89.2  & 88.1  & 87.9  \\
    ELECTRA &       & \textbf{90.0}  & \textbf{90.2}  & \textbf{88.4}  & \textbf{88.0}  \\
    \hline

    \hline
    \end{tabular}%
  }
  \caption{Main Results on MVIS and TOPv2 datasets.}
  \label{tab:main}%
\end{table}%

In main experiments, we merge all data samples, including single-domain and multi-domain, and feed ontology items from all domains as input. Thus the model needs to determine the specific domain(s) of the current utterance. We adopt the classic Sequence Labeling (SL) method~\cite{qin2021co} as the baseline. By treating frequent but unaligned ``(domain, intent, slot, value)'' quadruples as utterance-level labels, we add a multi-label \underline{CL}assi\underline{F}ier to tackle the unaligned slot value problem. From Table~\ref{tab:main}, we can observe that:

1) Compared to traditional methods SL and SL+CLF, sequence generation is more suitable for tackling the hierarchical semantic frames. The inherent limitation of SL-based methods makes it difficult to recover the semantic structure from a flattened sequence. 

2) Ontology copy mechanism is significant to ensure the consistency with pre-defined names of ontology items. This conclusion is verified not only with our customized pointer-generator BiRGAT decoder but also with the end-to-end BART~\cite{bart}
 model. 



\subsection{Ablation of BiRGAT Encoder}\label{sec:abl_enc}
\begin{table}[htbp]
  \centering
  \resizebox{0.49\textwidth}{!}{
    \begin{tabular}{c|c|c|c|cc}
    \hline

    \hline
    \textbf{Init} & \textbf{w/ OE} & \textbf{GNN} & \textbf{w/ DCA} & \textbf{MIVS} & \textbf{TOPv2}\\
    \hline\hline
    \multirow{6}{*}{SWV} & \XSolidBrush & \multirow{2}{*}{None} & \multirow{2}{*}{\XSolidBrush} & 83.3  & 82.6\\
          & {\large\checkmark} &       &       & 83.9  & 85.3 \\
\cdashline{2-6}          & \multirow{2}{*}{\large\checkmark} & \multirow{2}{*}{GAT} & \XSolidBrush & 84.1  & 85.3 \\
          &       &       & {\large\checkmark} & 84.8  & \textbf{85.9} \\
\cdashline{2-6}          & \multirow{2}{*}{\large\checkmark} & \multirow{2}{*}{RGAT} & \XSolidBrush & 84.7  & 85.2  \\
          &       &       & {\large\checkmark} & \textbf{85.6}  & \textbf{85.9}  \\
    \hline
    \multirow{6}{*}{BERT} & \XSolidBrush & \multirow{2}{*}{None} & \multirow{2}{*}{\XSolidBrush} & 88.4 & 86.1 \\
          & {\large\checkmark} &       &       &    89.1   &  87.6 \\
\cdashline{2-6}          & \multirow{2}{*}{\large\checkmark} & \multirow{2}{*}{GAT} & \XSolidBrush &  88.4   &  87.7 \\
          &       &       & {\large\checkmark} &    \textbf{89.4}   &  \textbf{87.8} \\
\cdashline{2-6}          & \multirow{2}{*}{\large\checkmark} & \multirow{2}{*}{RGAT} & \XSolidBrush &   89.1    & 87.3 \\
          &       &       & {\large\checkmark} &   89.3    &  \textbf{87.8}\\
    \hline

\hline
    \end{tabular}%
    }
  \caption{Ablation study on BiRGAT encoder. w/ OE: with ontology encoding; w/ DCA: with dual multi-head cross-attention.}
  \label{tab:abl_enc}%
\end{table}%

In this section, we study the contribution of each component in the BiRGAT encoder, including \underline{O}ntology \underline{E}ncoding module~(OE), relational features $\mathbf{z}_{ij}$ (GAT v.s. RGAT), and \underline{D}ual multi-head \underline{C}ross-\underline{A}ttention sub-module of GNN layer~(DCA). According to Table~\ref{tab:abl_enc}, we can summarize that:

1) Leveraging the text description of ontology items~(w/ OE) is effective in enriching the semantics of ontology embeddings. This observation is consistent for both two datasets and all settings.


2) Both structural~(GAT) and relational~(RGAT) encoding can boost the performance. Compared to TOPv2, our MIVS dataset seems to benefit more from the hierarchy of ontology items. It can be attributed to the fact that data samples in MIVS contain relatively more intents and exhibit more complicated output tree structures.

3) Although the separate encoding of question and ontology already achieves remarkable results, the integration of cross-segment attention still brings stable performance gains on both datasets.


\subsection{Transfer to More Intents}\label{sec:transfer_intent}
\begin{figure}[htbp]
    \centering
    \includegraphics[width=0.45\textwidth]{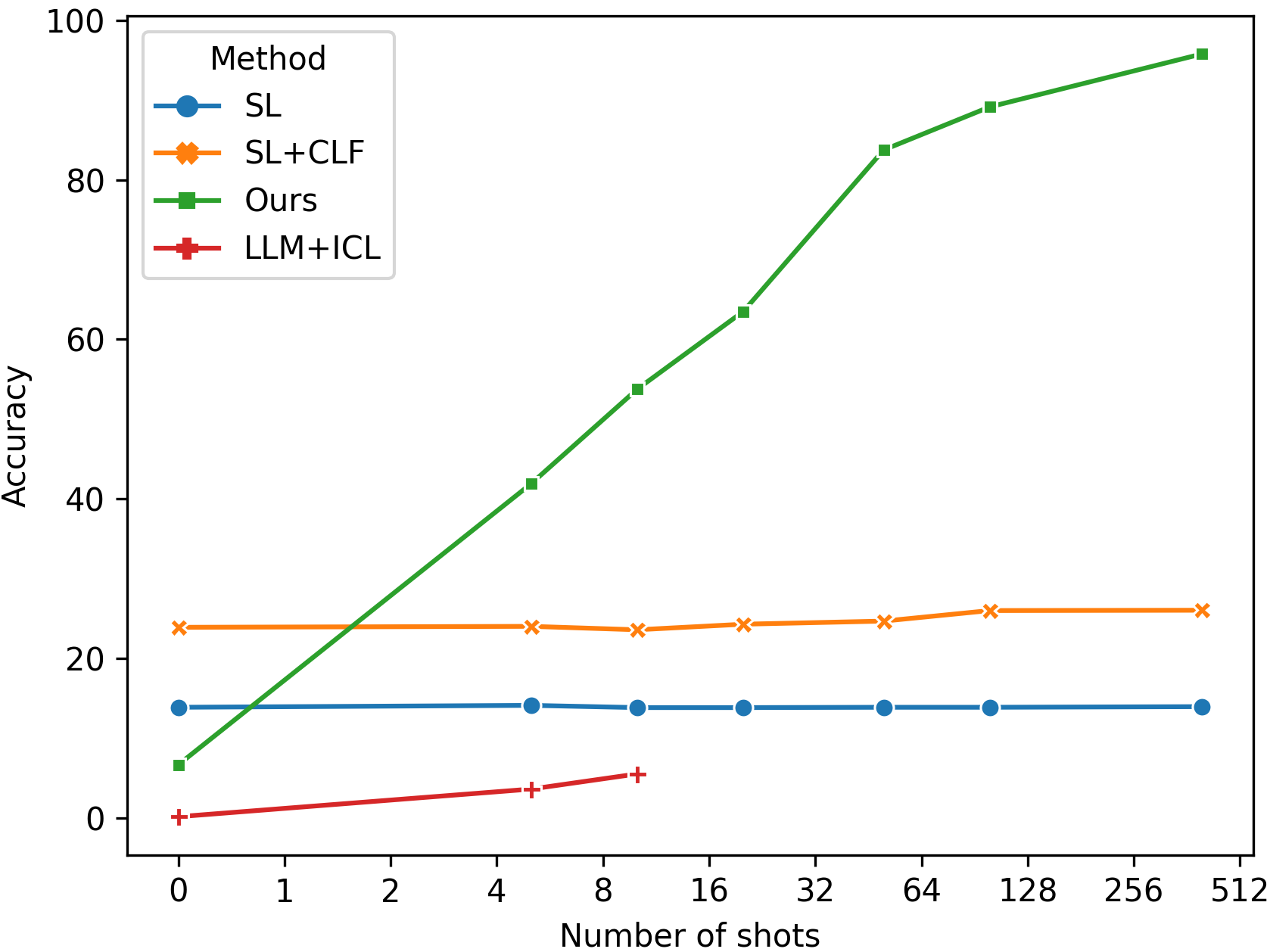}
    \caption{Few-shot learning experiments when transferring to more intents~($>3$) in domain ``{\it in-vehicle control}''. Due to the max token limit, prompts of LLM are truncated to at most $10$ exemplars.}
    \label{fig:intent}
\end{figure}

In this section, we explore the intent generalizability of current SLU models. Concretely, we train the model on data samples with the number of intents less or equal to $3$, but evaluate the model on examples containing more intents~($>3$). We can further fine-tune the model with a few samples containing $>3$ intents. 
We conduct experiments on domain ``{\it in-vehicle control}'' since it contains more intents on average. We also introduce a large language model~(LLM,~\cite{instructgpt}) baseline, i.e., \texttt{text-davinci-003} with in-context learning~(ICL) for comparison. From Figure~\ref{fig:intent} we can observe that:

1) Disappointingly, our method is less performant than SL-based methods in zero-shot settings. Through the case study, we find that most erroneous predictions merely contain $3$ intents and omit an entire intent sub-tree. We hypothesize that the generation-based method suffers from the problem of over-fitting the output length, while SL-based methods achieve better generalization of length variation. 

2) In few-shot settings, our method dramatically surpasses SL-based methods with merely $5$ samples. Moreover, SL-based methods attain limited improvements from fine-tuning. It can be explained that outputs with more intents usually present more complicated structures, which may be tough for SL-based methods to reconstruct. 

3) Despite exciting results in other fields, it is difficult for LLM to produce both semantically coherent and syntactically-valid output sequences with very few exemplars. 

\section{Conclusion}
In this work, we propose a large-scale multi-domain multi-intent dataset MIVS which is collected from a realistic in-vehicle dialogue system. Accordingly, we devise a BiRGAT model to incorporate the hierarchy of ontology items into the graph encoder and introduce a three-way copy mechanism to the decoder. Experiments on datasets MIVS and TOPv2 demonstrate the superiority of BiRGAT over various baselines.

\section{ACKNOWLEDGEMENTS}
This work is funded by the China NSFC Projects (92370206, U23B2057, 62106142 and 62120106006) and Shanghai Municipal Science and Technology Major Project (2021SHZDZX0102).

\bibliographystyle{IEEEbib}
\bibliography{custom}

\end{document}